\pdfoutput=1
\documentclass[11pt,letterpaper]{article}

\usepackage[margin=1in]{geometry}
\usepackage[T1]{fontenc}
\usepackage{lmodern}
\usepackage{microtype}
\usepackage{amsmath}
\usepackage{amssymb}
\usepackage{graphicx}
\usepackage{booktabs}
\usepackage{makecell}
\usepackage{multirow}
\usepackage[table]{xcolor}
\usepackage[numbers,sort&compress]{natbib}
\usepackage[font=small,labelfont=bf,labelsep=period]{caption}
\usepackage{placeins}
\PassOptionsToPackage{hyphens}{url}

\definecolor{linkblue}{rgb}{0.21,0.49,0.74}
\usepackage[pagebackref,breaklinks,colorlinks,allcolors=linkblue]{hyperref}
\hypersetup{
  pdftitle={AEGIS: Preserving Privacy of 3D Facial Avatars with Adversarial Perturbations},
  pdfauthor={Dawid Wolkiewicz, Anastasiya Pechko, Przemys\l{}aw Spurek, Piotr Syga}
}

\def\our{AEGIS}
\def\fullmname{Adversarial Evasion Gaussian Identity Shield}

\newcommand{\papertitle}{\our{}: Preserving Privacy of 3D Facial Avatars with Adversarial Perturbations}
\title{\papertitle}
\date{}

\author{
Dawid Wolkiewicz$^{1}$ \quad
Anastasiya Pechko$^{2}$ \quad
Przemys{\l}aw Spurek$^{2}$ \quad
Piotr Syga$^{1}$\\[2mm]
$^{1}$Wroc{\l}aw University of Science and Technology, Wroc{\l}aw, Poland\\
{\tt\small dawid.wolkiewicz@pwr.edu.pl, piotr.syga@pwr.edu.pl}\\[1mm]
$^{2}$Jagiellonian University, Krak\'ow, Poland\\
{\tt\small anastasiya.pechko@doctoral.uj.edu.pl, przemyslaw.spurek@uj.edu.pl}
}

\begin{document}
\maketitle
\begin{abstract}
The growing adoption of photorealistic 3D facial avatars, particularly those utilizing efficient 3D Gaussian Splatting representations, introduces new risks of online identity theft, especially in systems that rely on biometric authentication. While effective adversarial masking methods have been developed for 2D images, a~significant gap remains in achieving robust, viewpoint-consistent identity protection for dynamic 3D avatars. To address this, we present \our{}, the first privacy-preserving identity masking framework for 3D Gaussian avatars, concealing identity from automated face recognition while preserving the avatar's perceptual realism and animation usability. \our{} applies adversarial perturbations to the base (DC) color coefficients of the Gaussian primitives, guided by an ensemble of pre-trained face verification networks, embedding the protection directly in the 3D representation so that it applies across viewpoints and animations, without retraining or modifying the avatar's geometry. Evaluated against black-box recognizers, \our{} reduces face retrieval and verification accuracy to near-zero while maintaining high perceptual fidelity. It also preserves key facial attributes such as age, gender, and emotion, demonstrating strong privacy protection with minimal visual distortion. We further introduce a~viewpoint-robust leakage metric that quantifies residual identity leakage across an avatar's rendered viewpoints, which single-view evaluation cannot capture.
\end{abstract}

\begin{figure}[!h]
    \centering
    \includegraphics[width=\linewidth]{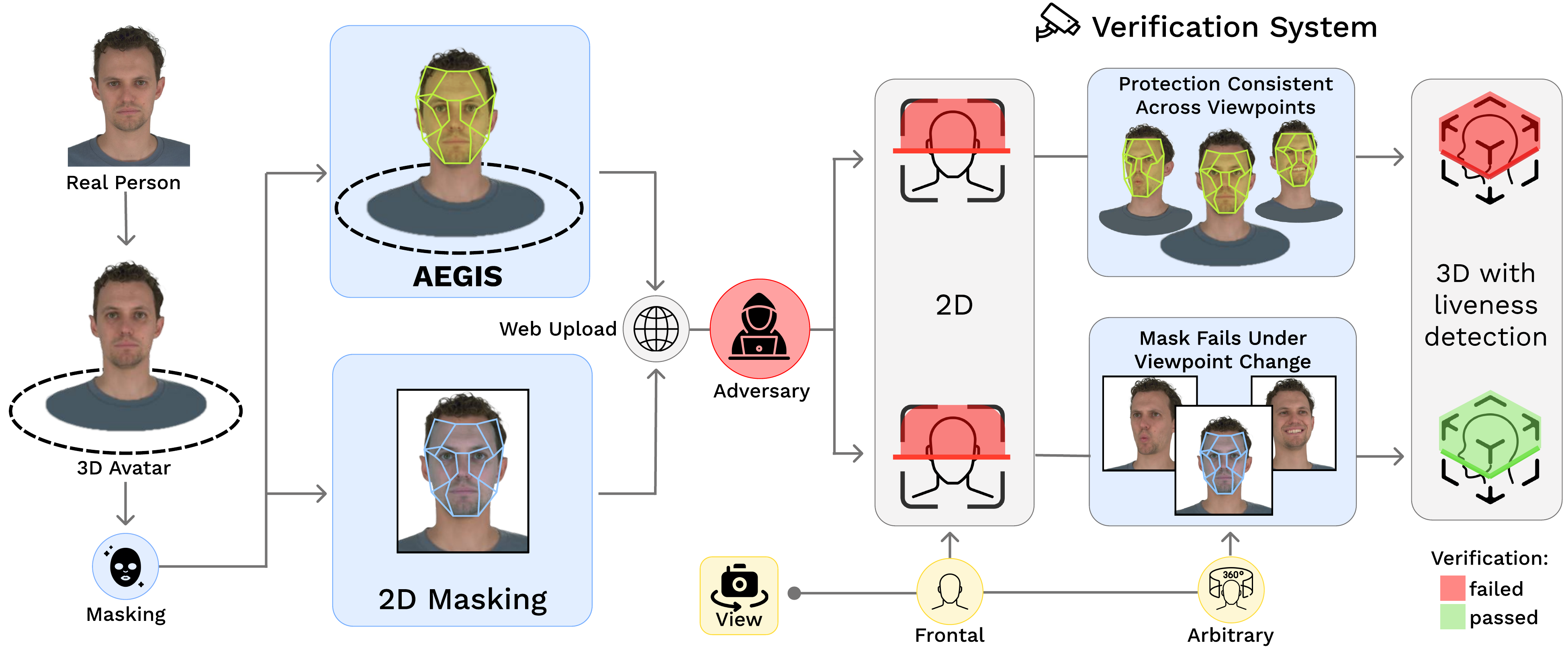}
    \caption{An adversary who obtains a~3D avatar can spoof RGB-based face verification, even under active liveness checks that prompt head rotation or expression changes and confirm identity from multiple viewpoints, because a~controllable photorealistic avatar renders consistently from any viewpoint in real time. A~2D mask is applied to each rendered frame rather than embedded in the asset, so the raw avatar can be re-rendered from any viewpoint with no protection present. \our{} instead embeds adversarial protection directly into the 3D representation, masking identity across viewpoints and poses so that the rendered avatar fails multi-view RGB verification}
    \label{fig:2dvs3d}
\end{figure}

\begin{figure}[tbp]
    \centering
    \includegraphics[width=\linewidth]{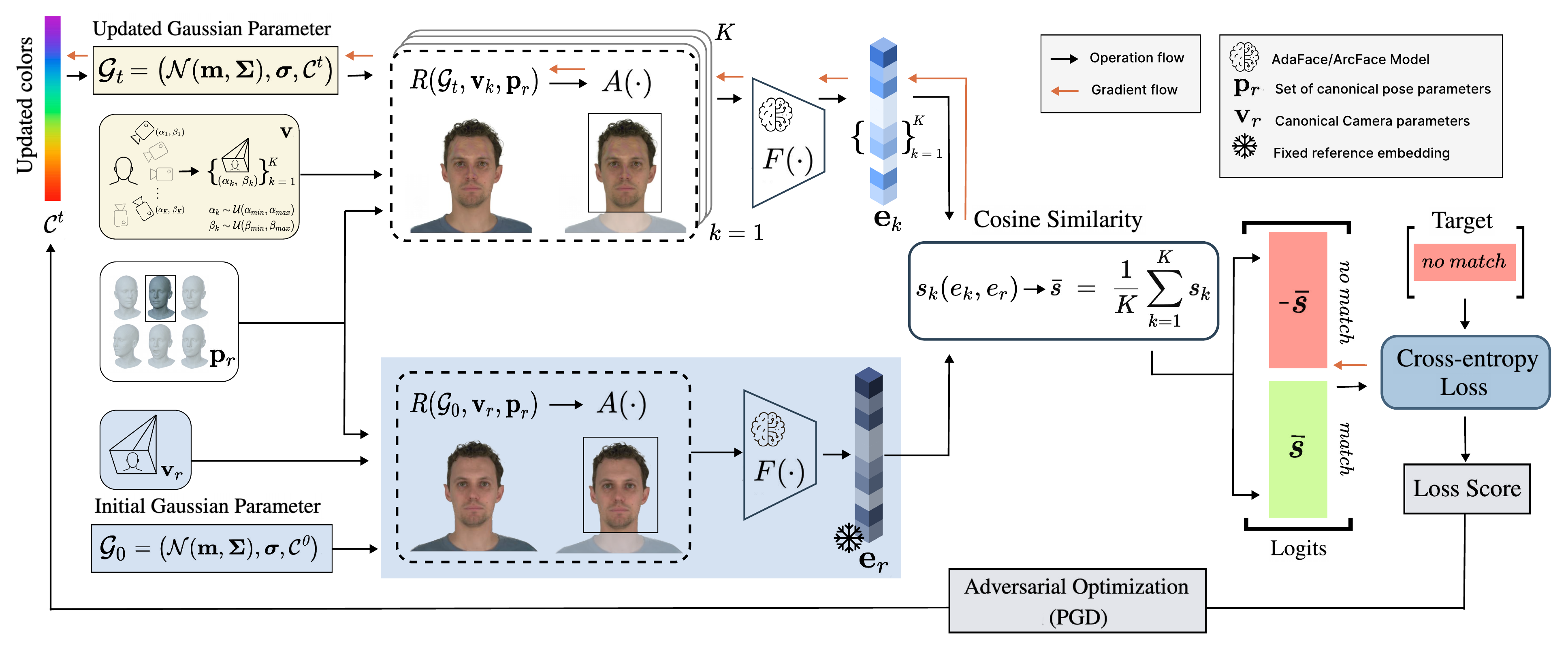}
    \caption{An overview of the \our{} identity masking pipeline, which adversarially optimizes the DC color coefficients $\mathcal{C}^t$ of a~3D Gaussian avatar to suppress its identity under an ensemble of face recognizers $F(\cdot)$. A~reference embedding $\mathbf{e}_r$ is computed once from the original avatar $\mathcal{G}_0$ at the canonical camera $\mathbf{v}_r$ and pose $\mathbf{p}_r$. Each PGD step samples camera parameters $\{ \mathbf{v}_k \}_{k=1}^K$ to cover diverse viewpoints, renders the current avatar $\mathcal{G}_t$ using $R(\cdot)$, aligns the crops with $A(\cdot)$, and extracts embeddings $\{\mathbf{e}_k\}$ whose mean cosine similarity $\bar{s}$ to $\mathbf{e}_r$, averaged over the sampled viewpoints and the ensemble, defines the match / no-match logits ($\bar{s}$ and $-\bar{s}$). A~cross-entropy loss targeting ``no match'' is backpropagated to update $\mathcal{C}^t$, yielding the masked avatar}
    \label{fig:pipeline}
\end{figure}

\section{Introduction}

With the rise of the metaverse~\cite{wang2023survey,tu2025we} and rapid photorealistic visualization, 3D facial avatars are increasingly prevalent online, especially those based on efficient 3D Gaussian Splatting (3DGS)~\cite{kerbl3Dgaussians}. Creating and widely sharing faithful, controllable digital personas raises a~critical security concern: identity theft via biometric extraction, especially severe for face verification and recognition authentication. While facial capture from online videos or calls is already a~privacy risk, the persistent, high-fidelity nature of a~controllable 3D avatar creates a~systemic vulnerability enabling large-scale spoofing and impersonation.

While biometric privacy for 2D imagery is well studied, methods such as Fawkes~\cite{fawkes} and LowKey~\cite{lowkey} use imperceptible pixel-space perturbations that shift user features in embedding space, and other work uses generative anonymization or adversarial editing. These 2D methods are ill-suited to 3D avatars: a~2D mask is applied to a~single rendered image rather than embedded in the avatar, so it leaves the source asset unprotected: an adversary can re-render the avatar from any viewpoint, or redistribute it, with no protection present (Fig.~\ref{fig:2dvs3d}). Real-time use is impractical as per-frame processing yields only ${\sim}1$ FPS with LowKey and ${\sim}0.56$ FPS with Fawkes on an NVIDIA RTX~4090 Laptop GPU at $512{\times}512$, far below ${\geq}30$ FPS streaming.

To address this, we introduce \our{} (\fullmname{}, see Fig.~\ref{fig:pipeline}), the first viewpoint-consistent identity obfuscation for 3D Gaussian avatars. It embeds protection into the 3D representation at optimization time, incurs no rendering overhead, and enables ${\sim}120$ FPS interactive playback on the same hardware.

Our goal is to evade automated face recognition while preserving the subject's likeness to human observers and the avatar's utility for legitimate use such as video calls and social VR. Unlike anonymization, which replaces the identity with a~different-looking one, a~protected avatar should still look like the user; only its machine-readable identity signature is suppressed. \our{} is a~post-training method that perturbs only the DC spherical-harmonics coefficients encoding the Gaussian primitives' base color, leaving geometry and higher-order view dependence unchanged. The obfuscation is learned via constrained Projected Gradient Descent (PGD)~\cite{pgd} against a~face verification objective using an ensemble of pre-trained face recognition models. Optimizing directly in avatar parameter space through differentiable rendering yields viewpoint-consistent, animation-stable, and perceptually coherent de-identification.

Below, we summarize the key contributions of the work:
\begin{itemize}
    \item We present \our{}, the first framework to embed identity protection directly into a~3D Gaussian avatar, perturbing only the DC spherical-harmonics color coefficients so that it persists across viewpoints and animations while geometry is preserved.

    \item On 60 avatars, \our{} reduces black-box verification and retrieval accuracy to near-zero, preserving perceptual fidelity and soft-trait consistency, outperforming seven 2D privacy baselines on the privacy--fidelity trade-off.

    \item We introduce a~viewpoint-robust leakage (VRL) metric that quantifies residual identity leakage across an avatar's rendered viewpoints, a~dimension not captured by conventional single-view evaluation.
\end{itemize}

\noindent\textbf{Code Availability.} Code is available at \url{https://github.com/Woleek/AEGIS}.

\section{Related Work}
Basic face anonymization methods, such as blurring or black rectangles, are sufficient for static or low-stakes scenarios but offer limited utility in dynamic contexts like social media or video calls and are vulnerable to reversal~\cite{9157139,Deblurringzhai2025}. Adversarial 2D protection methods, including Fawkes~\cite{fawkes} and LowKey~\cite{lowkey}, reduce recognition by shifting features or applying evasion filters, yet they often degrade under transfer to unseen recognizers and cannot ensure viewpoint-consistent anonymization for 3D avatars. Generative anonymization approaches, such as diffusion-based methods~\cite{anonsimple, salar2025weakdiffpure}, G$^2$Face~\cite{10644096}, latent code optimization~\cite{latentcode}, and adversarial makeup editing~\cite{diffusionmakeup}, preserve attributes while changing identity, but operate in 2D and cannot maintain a~persistent, controllable avatar or guarantee consistency under head motion. Controllable photorealistic head avatars, like GaussianAvatars~\cite{qian2024gaussianavatars}, enable high-fidelity reenactment across poses and expressions, but prior work has not addressed privacy against biometric verification. Our approach fills this gap by introducing asset-level adversarial masking on the DC coefficients of spherical harmonics, preserving geometry and view-dependent appearance while enforcing viewpoint-consistent identity suppression without assumptions on the adversary or introducing recovery risks. This provides a~practical solution for protecting identity in animated, photorealistic 3D avatars while maintaining perceptual realism and control.

\begin{figure}[tp]
    \centering
    \includegraphics[width=\linewidth]{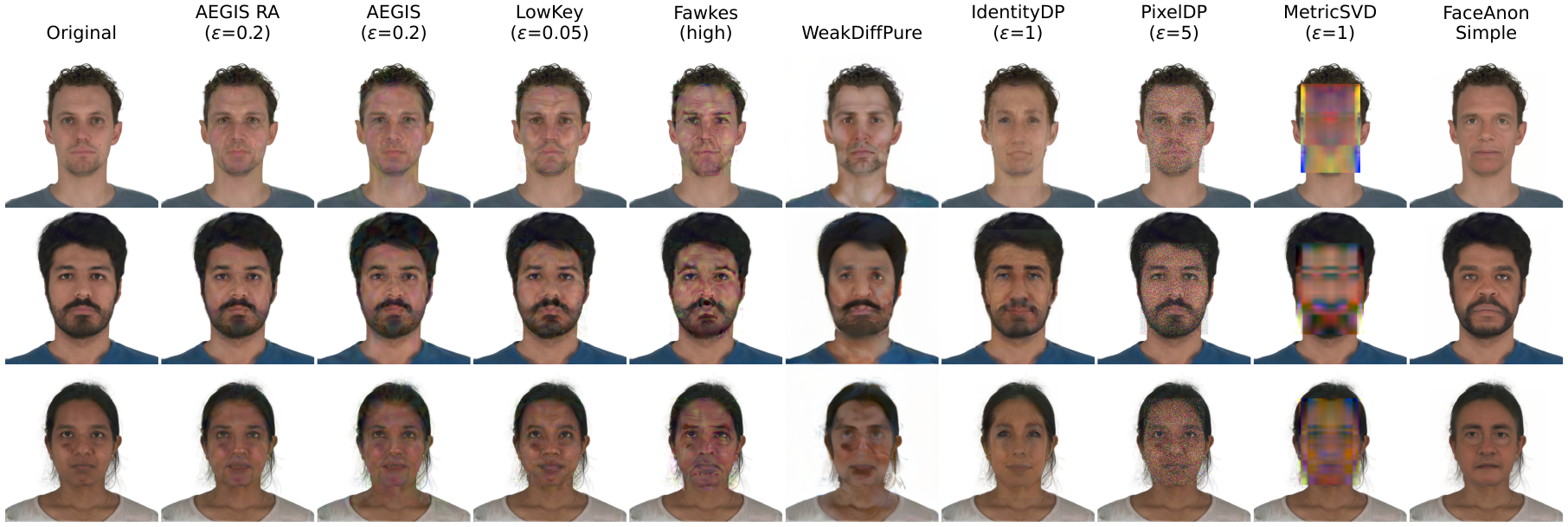}
    \caption{Visualization of example identity masks obtained using \our{} and reference 2D masking methods for three NeRSemble subjects}
    \label{fig:samples}

\end{figure}

\section{Method}
\label{sec:method}

The proposed avatar identity masking obfuscates identity-related features of a~3D Gaussian-based avatar~\cite{qian2024gaussianavatars} while preserving perceptual realism, by adversarially optimizing the DC coefficients, namely the zeroth-order spherical harmonics encoding each 3D Gaussian primitive's base, view-independent color~\cite{kerbl3Dgaussians}.

Our pipeline integrates three key components: (1) a~differentiable renderer for Gaussian avatars, (2) an ensemble of pre-trained face verification networks, each preceded by a~face-alignment preprocessing step, and (3) an adversarial optimization framework. An overview of the proposed method is presented in Fig.~\ref{fig:pipeline}, which illustrates the loop between rendering, face verification, and adversarial optimization over Gaussian color coefficients.

\paragraph{Threat Model}
We consider an adversary who obtains a~trained 3D avatar and renders it to spoof RGB-based face verification, including pipelines with active liveness checks that confirm identity from imagery captured across multiple viewpoints (e.g., prompting head rotation or expression changes), as used in remote onboarding and authentication. The adversary may render the avatar from arbitrary viewpoints; we optimize against a~surrogate ensemble of recognizers and evaluate on held-out ones (black-box transfer), assuming no access to real enrollment data.

\paragraph{Underlying Models}
The avatar uses the \textbf{GaussianAvatars} representation~\cite{qian2024gaussianavatars}, combining 3D Gaussian Splatting~\cite{kerbl3Dgaussians} for appearance with the FLAME parametric head model~\cite{FLAME:SiggraphAsia2017} for articulation. Our masking is applied to a~fully trained avatar and leaves the reconstruction process unchanged.

A~Gaussian-Splatting scene is a~set of $n$ anisotropic 3D primitives $\mathcal{G}=\left\{\left(\mathcal{N}(m_i,\Sigma_i),\sigma_i,C_i\right)\right\}_{i=1}^{n}$, each with a~mean $m_i$, covariance $\Sigma_i$, opacity $\sigma_i$, and a~color $C_i\in\mathbb{R}^{k\times3}$ encoded by $k$ spherical-harmonics (SH) coefficients per channel. Rendering differentiably alpha-composites the primitives in depth order, so the zeroth-order (DC) SH term, which sets each Gaussian's base view-independent color, directly determines the rendered appearance~\cite{Yifan_2019}. We perturb only these DC coefficients, leaving geometry and view-dependent terms intact.

FLAME~\cite{FLAME:SiggraphAsia2017} maps shape, pose, and expression parameters to a~3D head mesh through linear blend skinning. Beyond enabling animation, it provides semantic vertex correspondence, letting us identify the Gaussian primitives attached to specific facial regions (e.g., eyes, nose, skin), which we exploit for the region-adaptive variant described below. The DC coefficients of the selected Gaussians are assembled into a~tensor {$\mathcal{C}\in\mathbb{R}^{N\times3}$}, where $N$ is the number of selected Gaussians; $\mathcal{C}$ is the optimization variable in the adversarial masking process.

\paragraph{Facial Verification}
To quantify identity, we render the avatar from a~fixed viewpoint with a~consistent pose and extract its identity embedding. Let {$R(\mathcal{G}, \mathbf{v}, \mathbf{p})$} denote the differentiable rendering process that synthesizes a~2D image {$\mathcal{I}$} from the 3D Gaussian model {$\mathcal{G}$}, camera parameters {$\mathbf{v}$} defining viewpoint and projection, and pose parameters {$\mathbf{p}$} controlling the underlying FLAME model. Our optimization variable is the DC coefficients tensor {$\mathcal{C}$}, which defines the view-independent base color of the Gaussian primitives. We can therefore denote the parametrized model as {$\mathcal{G}_\mathcal{C}$}.

To process the rendered image {$\mathcal{I}$}, we apply a~face alignment function {$A(\cdot)$} that produces face crops of size $112\times112$ pixels, after pose normalization. The alignment is based on five facial landmarks predicted by the RetinaFace model~\cite{deng2020retinaface}, as in~\cite{deng2019arcface, kim2022adaface}. This operation compensates for variations in scale and in-plane rotation, yielding a~canonical input for the verification network. The aligned image {$A(\mathcal{I})$} is then passed to an ensemble of {$M$} pre-trained face verification networks {$\{F_m(\cdot)\}_{m=1}^{M}$}, each producing a~high-dimensional feature vector {$\mathbf{e}^{(m)} = F_m(A(\mathcal{I})) \in \mathbb{R}^{512}$}. We use IRSE50~\cite{hu2018squeeze}, IR152~\cite{deng2019arcface}, FaceNet~\cite{schroff2015facenet}, and MobileFace~\cite{chen2018mobilefacenets} as surrogates, following the standard black-box face-recognition evaluation set adopted in prior facial-privacy works~\cite{yang2021tipim, shamshad2023clip2protect, salar2025weakdiffpure}. Optimizing against several recognition systems simultaneously reduces reliance on any single model's decision boundary and improves transfer to unseen recognizers.

For adversarial optimization we use Expectation Over Transformation (EOT)~\cite{athalye2018synthesizing} to enforce viewpoint-robust masking. We define a~viewpoint distribution $\mathbf{T}$ over camera parameters $\mathbf{v}$ by uniformly sampling the pitch (x-rotation) and yaw (y-rotation) in $[-0.5,0.5]$ radians. At each step we sample $K$ viewpoints $\{\mathbf{v}_1,\ldots,\mathbf{v}_K\}\sim\mathbf{T}$ and use a~neutral reference pose $\mathbf{p}_r$. The identity embedding from surrogate $F_m$ for viewpoint $\mathbf{v}$ is a~differentiable function of the color tensor $\mathcal{C}$:
\begin{equation}
    \mathbf{e}_a^{(m)}(\mathcal{C}, \mathbf{v}) = F_m\big(A(R(\mathcal{G}_\mathcal{C}, \mathbf{v}, \mathbf{p}_r))\big)~,
\end{equation}
and each vector {$\mathbf{e}_a^{(m)}$} is L2-normalized to yield a~unit-length identity embedding.

For each surrogate, a~reference identity embedding {$\mathbf{e}_r^{(m)}$} is computed once from the original, unmasked avatar (with the original colors {$\mathcal{C}^0$}) rendered with identical pose and default frontal viewpoint, serving as the constant target to which the masked avatar's embedding is compared. The per-model cosine similarity quantifies the identity similarity (we use {$\cdot$} for scalar product):
\begin{equation}
    s_m(\mathcal{C}, \mathbf{v}) = \frac{\mathbf{e}_a^{(m)}(\mathcal{C}, \mathbf{v}) \cdot \mathbf{e}_r^{(m)}}{\|\mathbf{e}_a^{(m)}(\mathcal{C}, \mathbf{v})\| \|\mathbf{e}_r^{(m)}\|}~,
\end{equation}
and we average across the ensemble to obtain the aggregated identity similarity:
\begin{equation}
    \bar{s}(\mathcal{C}, \mathbf{v}) = \frac{1}{M}\sum_{m=1}^{M} s_m(\mathcal{C}, \mathbf{v})~.
\end{equation}

\paragraph{Adversarial Masking}
To hide the avatar identity, we adversarially optimize the DC coefficients tensor {$\mathcal{C}$}. We seek a~perturbed color tensor {$\mathcal{C}^\star$} that minimizes the expected ensemble-averaged identity similarity under the transformation distribution $\mathbf{T}$, thereby separating the masked avatar's embeddings from the reference embeddings across all surrogates. This constrained problem is formulated as:
\begin{equation}
\begin{split}
  \mathcal{C}^\star = \arg \min_{\mathcal{C}} \mathbf{E}_{\mathbf{v}\sim \mathbf{T}}\big[\bar{s}(\mathcal{C}, \mathbf{v})\big] \\
  \text{s.t.} \quad \|\mathcal{C} - \mathcal{C}^0\|_\infty \le \epsilon~.
\end{split}
\end{equation}
The constraint employs the $\ell_\infty$ norm, $\|\mathcal{C} - \mathcal{C}^0\|_\infty = \max_i |\mathcal{C}_i - \mathcal{C}^0_i|$, restricting the optimization to a~bounded hypercube around the original color tensor and preventing drastic color shifts or structural artifacts. The hyperparameter $\epsilon$ defines the perturbation budget, i.e., the maximum allowed per-channel deviation, bounded by the dynamic range of the original tensor, {$0 \le \epsilon \le \max(\mathcal{C}^0) - \min(\mathcal{C}^0)$}. Its value is selected empirically to balance privacy and perceptual realism.

We solve this with PGD, using a~cross-entropy loss $\mathcal{L}$ on the aggregated similarity $\bar{s} = \bar{s}(\mathcal{C}, \mathbf{v})$. For a~single viewpoint $\mathbf{v}$, the loss is:
\begin{equation}
    \mathcal{L}(\mathcal{C}, \mathbf{v}) = \log(1 + e^{-2\bar{s}\lambda})~,
\end{equation}
with logits $[-\bar{s}\lambda, \bar{s}\lambda]$ and a~scaling constant $\lambda$. The expected loss under the transformation distribution is:
\begin{equation}
    \mathcal{L}_{EOT}(\mathcal{C}) = \mathbf{E}_{\mathbf{v}\sim \mathbf{T}}\big[\mathcal{L}(\mathcal{C}, \mathbf{v})\big].
\end{equation}
Maximizing this loss corresponds to minimizing the identity similarity. The gradient $\nabla_{\mathcal{C}}\mathcal{L}_{EOT}$ thus provides the direction in color space that most effectively increases dissimilarity from the reference identity.

At each iteration $t$, PGD performs (1) a~gradient ascent step to maximize the loss, followed by (2) a~projection step enforcing the $\ell_\infty$ constraint. The gradient is estimated stochastically over $K$ sampled viewpoints:
\begin{equation}
    \nabla_{\mathcal{C}}\mathcal{L}_{EOT}(\mathcal{C}^{(t)}) \approx \frac{1}{K} \sum_{i=1}^{K} \nabla_{\mathcal{C}} \mathcal{L}(\mathcal{C}^{(t)}, \mathbf{v}_i),
\end{equation}
and the update rule is:
\begin{equation}
    \mathcal{C}^{(t+1)} = \phi_{\mathcal{C}^0, \epsilon}\left(
    \mathcal{C}^{(t)} + \alpha \cdot \text{sign}\big(\nabla_{\mathcal{C}}\mathcal{L}_{EOT}(\mathcal{C}^{(t)})\big)
    \right),
\end{equation}
where $\alpha$ is the step size controlling the update magnitude, set relative to the perturbation budget $\epsilon$. The $\text{sign}$ function specifies the optimal step direction within the $\ell_\infty$-norm ball. The projection operator {$\phi_{\mathcal{C}^0, \epsilon}(\cdot)$} enforces the constraint by clipping each element, $\phi_{\mathcal{C}^0,\epsilon}(\cdot)=\text{clip}(\cdot,\mathcal{C}^0-\epsilon,\mathcal{C}^0+\epsilon)$. After $T_{\max}$ iterations, the final optimized tensor {$\mathcal{C}^\star = \mathcal{C}^{(T_{\mathrm{max}})}$} replaces the original DC coefficients in the avatar model, yielding a~de-identified avatar representation that persistently conceals identity.

\paragraph{Region-Adaptive Perturbation Budget}
To reduce visual artifacts in perceptually-sensitive regions while maintaining strong de-identification in identity-critical areas, we further propose a~region-adaptive variant of \our{} (denoted \our{} RA). Rather than applying a~uniform $\epsilon$ across all Gaussians, we leverage the FLAME mesh topology to assign region-specific perturbation budget multipliers. Identity-critical areas receive higher multipliers (eyes $1.3\times$, nose $1.2\times$) to concentrate the modifications where face recognition models focus most, while smooth, perceptually sensitive regions receive lower multipliers (skin $0.7\times$, forehead $0.6\times$, scalp $0.4\times$) to minimize visible artifacts. These multipliers were determined empirically by evaluating the privacy-quality trade-off across a~set of candidate configurations (Appendix~\ref{sect:ra_sweep}). The resulting spatially-varying budget improves the privacy-quality trade-off over the uniform-$\epsilon$ baseline.

\paragraph{Implementation Details}
All experiments use $T_{\mathrm{max}}=200$ PGD steps with step size $\alpha=(0.01/0.3)\epsilon$, $K=5$ EOT viewpoints per step, and loss scaling $\lambda=10$. We optimize against three of the four surrogates using mean cross-model aggregation and evaluate on the held-out recognizer, reporting leave-one-out results over all four choices. Masking covers the full avatar head region unless otherwise specified. Experiments run on a~single NVIDIA RTX~4090 Laptop GPU (16\,GB).

\section{Evaluation}
\label{sec:evaluation}

To systematically assess the proposed avatar identity masking framework, we evaluate the masked avatars from four complementary perspectives: (1) identity retrieval, (2) face verification, (3) viewpoint-robust leakage, and (4) visual utility preservation.

Our evaluation method extends~\cite{wilson2025towards}, unifying privacy and utility assessment across 2D and 3D settings by combining rendered avatar datasets with established 2D face databases, rather than evaluating them on separate benchmarks.

In the privacy-preserving evaluations, existing large-scale 2D face datasets are employed to emulate a~real-world verification system with multiple enrolled identities. Specifically, we use CelebA~\cite{liu2015faceattributes}, a~large-scale dataset of over 200k celebrity faces serving as a~diverse retrieval gallery, and Labeled Faces in the Wild (LFW)~\cite{huang2008labeled}, a~benchmark dataset of unconstrained face photographs for threshold calibration and verification testing. Additionally, we construct a~reference dataset of 60 subjects from the NeRSemble dataset~\cite{kirschstein2023nersemble} used to train Gaussian avatars, introducing diversity in facial geometry, skin tone, and demographic representation. For each subject, frames from 11 different motion-capture recordings are available, each capturing distinct emotions and expressions.

To match face verification models, we keep only near-frontal images. Frontal yaw, pitch, and roll thresholds are set by rendering avatars with default camera parameters and estimating pose from RetinaFace landmarks, then slightly relaxing them using real-image deviations. Per subject we sample 22 frames, 2~per expression category, matching the CelebA mean images per identity (19.91), and denote this dataset NeRSembleGT. All experiments evaluate masking on 2D avatar renderings since metrics and verification models require 2D inputs.

\paragraph{Identity Retrieval}
The rank-$k$ identification evaluation measures how easily masked avatars can be linked back to their original identity among a~gallery of images. This experiment reflects a~realistic retrieval scenario in which a~privacy-preserving system must ensure that masked faces are sufficiently displaced in the embedding space, preventing re-identification.

For each masked render (query), we extract a~face embedding {$\mathbf{e}_q$} using a~face recognition model held out from the masking ensemble (Sec.~\ref{sec:method}), measuring black-box transferability across recognition systems. The gallery {$\{\mathbf{e}_g\}$} holds one template per identity: each of the 60 NeRSemble subjects is represented by the average embedding over its NeRSembleGT frames, augmented with 10{,}177 distractor identities from CelebA. Cosine distance is used to rank all gallery embeddings by similarity to the query:
\begin{equation}
    d_{i} = 1 - \frac{\mathbf{e}_{q} \cdot \mathbf{e}_{g_{i}}}{\|\mathbf{e}_{q}\| \|\mathbf{e}_{g_{i}}\|}~.
\end{equation}

We report the rank-$k$ retrieval rate (Rank-1 and Rank-50) for {$k\in\{1, 50\}$}:
\begin{equation}
    \text{Rank-}k = \frac{1}{N} \sum_{i=1}^{N} \mathbf{I}[\text{rank}(i) \leq k]~,
\end{equation}
where {$N$} is the number of queries and {$\mathbf{I}$} is the indicator function; lower values indicate stronger de-identification. Rank-50 is the more lenient retrieval criterion and thus the more demanding test of the masking.

\paragraph{Face Verification}
While rank-$k$ retrieval evaluates resistance to identification, face verification assesses whether the masked avatar can still be accepted as the same individual under standard biometric protocols.

The per-model decision threshold {$\tau$} for cosine similarity is calibrated on the LFW dataset~\cite{huang2008labeled} from sampled positive and negative pairs, using the False Accept Rate (FAR) and False Reject Rate (FRR) as functions of {$\tau$}. Following standard verification practice, we fix the operating point at a~low false-accept budget of {$\text{FAR}=10^{-3}$}, setting {$\tau$} to the lowest similarity threshold satisfying this constraint, i.e., the highest True Accept Rate ({$\text{TAR}=1-\text{FRR}$}) at that FAR. A~separate {$\tau$} is calibrated for each recognition model.

We evaluate masked avatars against their original NeRSembleGT references using a~fixed per-model threshold. For each {$(\mathbf{e}_a,\mathbf{e}_r)$}, cosine similarity {$s(\mathbf{e}_a,\mathbf{e}_r)$} determines a~\textit{match} if {$s\geq\tau$} and a~\textit{no-match} otherwise. The verification rate is the fraction of masked avatars matched to the originals (lower rates indicate stronger privacy).

\paragraph{Viewpoint-Robust Leakage}
Re-identification and verification rates, the standard privacy metrics in facial privacy protection, are conventionally measured on a~single near-frontal view. A~3D avatar, however, exposes its entire rendered viewpoint range to the attacker: an adversary who obtains the asset can render it and select the angle at which the masking is weakest. Single-view metrics inherited from the 2D setting therefore do not bound the identity an attacker can recover, and a~defense that protects a~3D asset must remain robust as the avatar is rotated. We evaluate protection over the rendered viewpoint cone within which the face remains detectable.

For each subject $u\in\mathcal{U}$ we render a~$9\times9$ grid of viewpoints $v\in\mathcal{V}$ (orbit angles sampled in $[-0.5,0.5]$ radians), embed every rendered view with a~held-out recognizer $F$, and compute the cosine similarity
\begin{equation}
    \sigma_u(v) = \cos\big(F(A(R(\mathcal{G}_{\mathcal{C}^\star}, v, \mathbf{p}_r))),\, \mathbf{g}_u\big)
\end{equation}
to the subject's identity template $\mathbf{g}_u$, the mean L2-normalized gallery embedding used in the retrieval protocol. We summarize leakage as the survival curve over all detected $(u,v)$ pairs,
\begin{equation}
    L(\tau) = \Pr_{u,v}\big[\sigma_u(v) \ge \tau\big],
\end{equation}
the re-identification rate as a~function of the decision threshold $\tau$. At each recognizer's verification threshold $\tau^\star$, $L(\tau^\star)$ gives the leakage rate at the system's operating point, while
\begin{equation}
    \text{VRL-AUC} = \frac{1}{1-\tau^\star}\int_{\tau^\star}^{1} L(\tau)\,d\tau \;\in\; [0,1]
\end{equation}
summarizes the entire re-identifiable region $[\tau^\star,1]$ as the normalized leakage mass above threshold (lower is better), reflecting how far similarities extend into the recognizable zone rather than their value at a~single threshold; we integrate only above $\tau^\star$ since the full-range area reduces to a~monotone restatement of mean similarity. Finally, because the attacker may choose the most revealing viewpoint, we report a~worst-case guarantee,
\begin{equation}
    \text{WC} = \frac{1}{|\mathcal{U}|}\sum_{u\in\mathcal{U}} \mathbf{1}\big[\textstyle\max_{v\in\mathcal{V}}\,\sigma_u(v) \ge \tau^\star\big],
\end{equation}
the fraction of subjects re-identifiable from at least one rendered viewpoint, a~quantity no single-image 2D protocol can express.

\paragraph{Utility Preservation}
We assess utility preservation of the masked avatars relative to their unaltered counterparts using low-level fidelity and high-level semantic metrics.

For each original/masked render pair we compute SSIM (structural consistency) and PSNR (pixel-level fidelity), after aligning and cropping both images to the RetinaFace bounding box, and average over all pairs.

To evaluate soft-trait preservation, we use the DeepFace framework~\cite{serengil2024lightface} to predict facial attributes from original and masked render pairs. We compute four metrics: the emotion match rate (agreement in predicted dominant emotion), gender match rate (agreement in predicted gender), race match rate (agreement in predicted racial category), and age difference (the absolute difference between predicted ages). High SSIM/PSNR and semantic consistency indicate that the masking procedure preserves perceptual realism and usability, while low identification and verification scores confirm effective identity protection.

\section{Experiments}
\label{sec:experiments}

\begin{table}[tbp]
\centering
\renewcommand{\arraystretch}{1}
\caption{Privacy evaluation on the 60 NeRSemble subjects. \textbf{Verification} is the per-model rate at which the masked avatar still matches the real identity (TAR@FAR$=10^{-3}$, single reference); every scoring recognizer is unseen during masking. \textbf{Retrieval} (Rank-1/Rank-50) is pooled as mean\,$\pm$\,std over the four black-box recognizers, against a~gallery of 60 identity templates plus 10{,}177 CelebA distractors. \textbf{Viewpoint robustness} (VRL-AUC, leakage $L(\tau^\star)$, worst-case WC) applies only to 3D representations; 2D methods have no persistent avatar. Lower is more private; \textbf{bold} marks the best per column and {\setlength{\fboxsep}{0.5pt}\colorbox{gray!12}{shading}} the recommended operating point}
\label{tab:privacy}
{\fontsize{7pt}{9pt}\selectfont
\resizebox{\textwidth}{!}{
\begin{tabular}{lrrrrc|cc|ccc}
\hline
 & \multicolumn{5}{c|}{\textbf{Verification (\%) $\downarrow$}} & \multicolumn{2}{c|}{\textbf{Retrieval (\%) $\downarrow$}} & \multicolumn{3}{c}{\textbf{Viewpoint robustness $\downarrow$}} \\
\bf Method & \bf IRSE50 & \bf IR152 & \bf FaceNet & \bf Mobile & \bf Avg & \bf Rank-1 & \bf Rank-50 & \bf VRL-AUC & \boldmath$L(\tau^\star)$ & \bf WC \\
\hline
Avatar renders & 100.00 & 100.00 & 100.00 & 100.00 & 100.00 & 100.00 & 100.00 & $0.693 \pm 0.056$ & $0.99 \pm 0.01$ & $1.00 \pm 0.00$ \\
\hline
\rowcolor{gray!12}\our{} RA ($\epsilon{=}0.2$) & \textbf{0.00} & 3.33 & 1.67 & \textbf{0.00} & $1.25 \pm 1.38$ & $0.83 \pm 1.44$ & $6.25 \pm 7.49$ & $0.016 \pm 0.012$ & $0.10 \pm 0.07$ & $0.36 \pm 0.16$ \\
\our{} base ($\epsilon{=}0.1$) & 18.33 & 51.67 & 26.67 & 11.67 & $27.08 \pm 15.16$ & $24.58 \pm 11.27$ & $61.25 \pm 10.23$ & $0.105 \pm 0.032$ & $0.59 \pm 0.10$ & $0.98 \pm 0.01$ \\
\our{} base ($\epsilon{=}0.2$) & \textbf{0.00} & \textbf{0.00} & \textbf{0.00} & \textbf{0.00} & \textbf{0.00} & \textbf{0.00} & $0.42 \pm 0.72$ & $0.002 \pm 0.002$ & $0.02 \pm 0.02$ & $0.18 \pm 0.14$ \\
\our{} base ($\epsilon{=}0.3$) & \textbf{0.00} & \textbf{0.00} & \textbf{0.00} & \textbf{0.00} & \textbf{0.00} & \textbf{0.00} & \textbf{0.00} & \textbf{0.000} & \textbf{0.00} & \textbf{0.03} \\
\hline
LowKey ($\epsilon{=}0.05$) & 20.00 & 5.00 & 20.00 & 5.00 & $12.50 \pm 7.50$ & $13.75 \pm 6.71$ & $39.58 \pm 13.76$ & -- & -- & -- \\
Fawkes (high) & 11.67 & \textbf{0.00} & 6.67 & 3.33 & $5.42 \pm 4.31$ & $5.42 \pm 3.80$ & $20.42 \pm 11.87$ & -- & -- & -- \\
WeakDiffPure & 3.75 & 3.75 & 5.42 & 2.08 & $3.75 \pm 1.18$ & $4.23 \pm 1.42$ & $13.92 \pm 4.39$ & -- & -- & -- \\
IdentityDP ($\epsilon{=}1$) & 1.67 & \textbf{0.00} & 3.33 & 1.67 & $1.67 \pm 1.18$ & $1.67 \pm 1.67$ & $17.08 \pm 10.30$ & -- & -- & -- \\
IdentityDP ($\epsilon{=}100$) & 11.67 & 1.67 & 13.33 & 11.67 & $9.58 \pm 4.62$ & $11.25 \pm 5.70$ & $34.58 \pm 18.27$ & -- & -- & -- \\
PixelDP ($\epsilon{=}5$) & 41.67 & 15.00 & 53.33 & 1.67 & $27.92 \pm 20.56$ & $31.67 \pm 25.85$ & $52.50 \pm 36.49$ & -- & -- & -- \\
PixelDP ($\epsilon{=}20$) & 100.00 & 100.00 & 100.00 & 100.00 & $100.00$ & $100.00$ & $100.00$ & -- & -- & -- \\
MetricSVD ($\epsilon{=}1$) & 8.33 & 1.67 & 3.33 & \textbf{0.00} & $3.33 \pm 3.12$ & $3.49 \pm 3.49$ & $25.58 \pm 17.48$ & -- & -- & -- \\
MetricSVD ($\epsilon{=}20$) & 68.33 & 51.67 & 43.33 & 33.33 & $49.17 \pm 12.83$ & $39.58 \pm 19.70$ & $84.17 \pm 8.29$ & -- & -- & -- \\
FaceAnonSimple & 6.67 & 3.33 & 3.33 & 6.67 & $5.00 \pm 1.67$ & $1.67 \pm 1.67$ & $22.50 \pm 12.61$ & -- & -- & -- \\
\hline
\end{tabular}
}
}
\end{table}

\begin{table}[tbp]
\centering
\renewcommand{\arraystretch}{1}
\caption{Utility evaluation on the 60 NeRSemble subjects. SSIM, PSNR, and FID are reported as mean\,$\pm$\,std; per-subject attribute-match rates (Emotion, Gender, Race) and the age difference are reported as means, as their per-subject standard deviations are dominated by the binary/quantized measurement structure. SSIM, PSNR, and age are pooled over subjects, while FID and attribute rates are computed over the evaluation folds. \textbf{Bold} marks the best value per column (excluding the unmasked reference) and {\setlength{\fboxsep}{0.5pt}\colorbox{gray!12}{shading}} marks the recommended operating point. Methods with strong attribute retention but no privacy (e.g., PixelDP $\epsilon{=}20$, cf.\ Tab.~\ref{tab:privacy}) leave the identity essentially intact}
\label{tab:utility}
{\footnotesize\setlength{\tabcolsep}{4pt}
\begin{tabular}{lccccccc}
\hline
\bf Method & \bf SSIM $\uparrow$ & \makecell{\textbf{PSNR}\\\textbf{(dB) $\uparrow$}} & \bf FID $\downarrow$ & \makecell{\textbf{Emotion}\\\textbf{(\%) $\uparrow$}} & \makecell{\textbf{Gender}\\\textbf{(\%) $\uparrow$}} & \makecell{\textbf{Race}\\\textbf{(\%) $\uparrow$}} & \makecell{\textbf{Age}\\\textbf{(diff.) $\downarrow$}} \\
\hline
Avatar renders & $1$ & $\infty$ & $0$ & $100$ & $100$ & $100$ & $0$ \\
\hline
\rowcolor{gray!12}\our{} RA ($\epsilon{=}0.2$) & $0.9347 \pm 0.0054$ & $32.66 \pm 0.32$ & $29.62 \pm 3.87$ & $80.25$ & $91.00$ & $45.75$ & $3.63$ \\
\our{} base ($\epsilon{=}0.1$) & $\mathbf{0.9586 \pm 0.0037}$ & $\mathbf{35.03 \pm 0.23}$ & $\mathbf{22.52 \pm 2.59}$ & $\mathbf{87.50}$ & $91.25$ & $57.25$ & $2.87$ \\
\our{} base ($\epsilon{=}0.2$) & $0.8837 \pm 0.0080$ & $29.62 \pm 0.31$ & $54.43 \pm 6.58$ & $74.50$ & $91.75$ & $32.75$ & $3.93$ \\
\our{} base ($\epsilon{=}0.3$) & $0.8124 \pm 0.0116$ & $26.78 \pm 0.40$ & $75.96 \pm 7.71$ & $73.25$ & $92.00$ & $25.75$ & $5.05$ \\
\hline
LowKey ($\epsilon{=}0.05$) & $0.8528 \pm 0.0265$ & $30.65 \pm 0.81$ & $45.73 \pm 9.33$ & $78.25$ & $90.25$ & $39.00$ & $4.54$ \\
Fawkes (high) & $0.8694 \pm 0.0113$ & $27.22 \pm 1.18$ & $79.61 \pm 5.07$ & $69.25$ & $90.25$ & $36.50$ & $6.27$ \\
WeakDiffPure & $0.8027 \pm 0.0301$ & $25.08 \pm 0.97$ & $62.37 \pm 5.76$ & $57.00$ & $95.00$ & $47.00$ & $7.55$ \\
IdentityDP ($\epsilon{=}1$) & $0.8141 \pm 0.0281$ & $26.76 \pm 0.97$ & $35.99 \pm 4.23$ & $61.00$ & $87.00$ & $40.00$ & $4.12$ \\
IdentityDP ($\epsilon{=}100$) & $0.8180 \pm 0.0287$ & $27.00 \pm 0.96$ & $34.67 \pm 3.32$ & $65.00$ & $79.00$ & $47.00$ & $3.03$ \\
PixelDP ($\epsilon{=}5$) & $0.0440 \pm 0.0051$ & $12.71 \pm 0.10$ & $303.93 \pm 7.13$ & $43.00$ & $90.00$ & $64.00$ & $7.65$ \\
PixelDP ($\epsilon{=}20$) & $0.2715 \pm 0.0167$ & $23.11 \pm 0.06$ & $64.29 \pm 0.94$ & $79.00$ & $\mathbf{99.00}$ & $\mathbf{94.00}$ & $\mathbf{2.00}$ \\
MetricSVD ($\epsilon{=}1$) & $0.6713 \pm 0.0581$ & $17.41 \pm 2.09$ & $254.64 \pm 4.68$ & $39.00$ & $90.00$ & $46.00$ & $10.40$ \\
MetricSVD ($\epsilon{=}20$) & $0.8334 \pm 0.0278$ & $26.83 \pm 1.43$ & $134.55 \pm 17.47$ & $65.00$ & $85.00$ & $56.00$ & $3.13$ \\
FaceAnonSimple & $0.8078 \pm 0.0311$ & $26.08 \pm 1.37$ & $30.98 \pm 3.94$ & $64.00$ & $91.00$ & $46.00$ & $4.13$ \\
\hline
\end{tabular}
}
\end{table}

We evaluate \our{} across privacy budgets and masking configurations on the 60 NeRSemble avatars described in Sec.~\ref{sec:evaluation}, under the surrogate-ensemble leave-one-out transfer protocol. Fig.~\ref{fig:samples} shows example renderings and qualitative comparisons to 2D masking.

We apply \our{} with privacy budgets $\epsilon\in\{0.1,0.2,0.3\}$ for the uniform variant and $\epsilon=0.2$ for the region-adaptive variant (\our{} RA), optimizing against the surrogate ensemble of Sec.~\ref{sec:method} and evaluating each held-out recognizer in turn. We benchmark against differentially private anonymizers, IdentityDP~\cite{wen2024differential}, PixelDP~\cite{fan2018image}, and MetricSVD~\cite{fan2019practical}, and adversarial or generative 2D methods, Fawkes~\cite{fawkes}, LowKey~\cite{lowkey}, WeakDiffPure~\cite{salar2025weakdiffpure}, and FaceAnonSimple~\cite{anonsimple}. Tables~\ref{tab:privacy} and~\ref{tab:utility} report privacy and utility, respectively. Because the 2D baselines protect only individual rendered frames rather than the avatar asset (Fig.~\ref{fig:2dvs3d}), the viewpoint-robustness metrics of Table~\ref{tab:privacy} apply only to \our{}, and the 2D methods contextualize per-render privacy rather than equivalent 3D protection.

Table~\ref{tab:privacy} reports privacy under black-box transfer. Uniform \our{} suppresses frontal recognition completely at $\epsilon=0.2$, where verification and Rank-1 retrieval reach $0\%$ against every held-out recognizer; $\epsilon=0.1$ is too weak and leaves identity transferable, while $\epsilon=0.3$ further reduces residual Rank-50 and viewpoint leakage at an added fidelity cost (Tab.~\ref{tab:utility}). \our{} RA ($\epsilon=0.2$) attains comparable suppression ($1.25\%$ verification, $0.83\%$ Rank-1) at substantially higher fidelity, and is our recommended operating point. No baseline matches \our{} RA across both metrics. IdentityDP ($\epsilon{=}1$), the closest in verification rate ($1.67\%$ vs.\ $1.25\%$), retains higher retrieval leakage (Rank-50 $17.08\%$ vs.\ $6.25\%$).

Table~\ref{tab:privacy} reveals a~viewpoint-dependent privacy dimension inaccessible to frontal, per-frame evaluation and inapplicable to 2D methods. The unmasked avatar remains re-identifiable across nearly all viewpoints in its cone ($L(\tau^\star){=}0.99$, WC$=1.00$), while uniform masking at $\epsilon=0.3$ nearly eliminates this leakage (WC$=0.03$). Despite leading in frontal and average metrics, \our{} RA exhibits higher worst-case leakage (WC$=0.36$), as concentrating the budget on identity-critical regions allows residual identity to emerge at oblique angles, exposing a~direct privacy--utility trade-off.

\begin{figure}[tbp]
    \centering
    \includegraphics[width=0.6\linewidth]{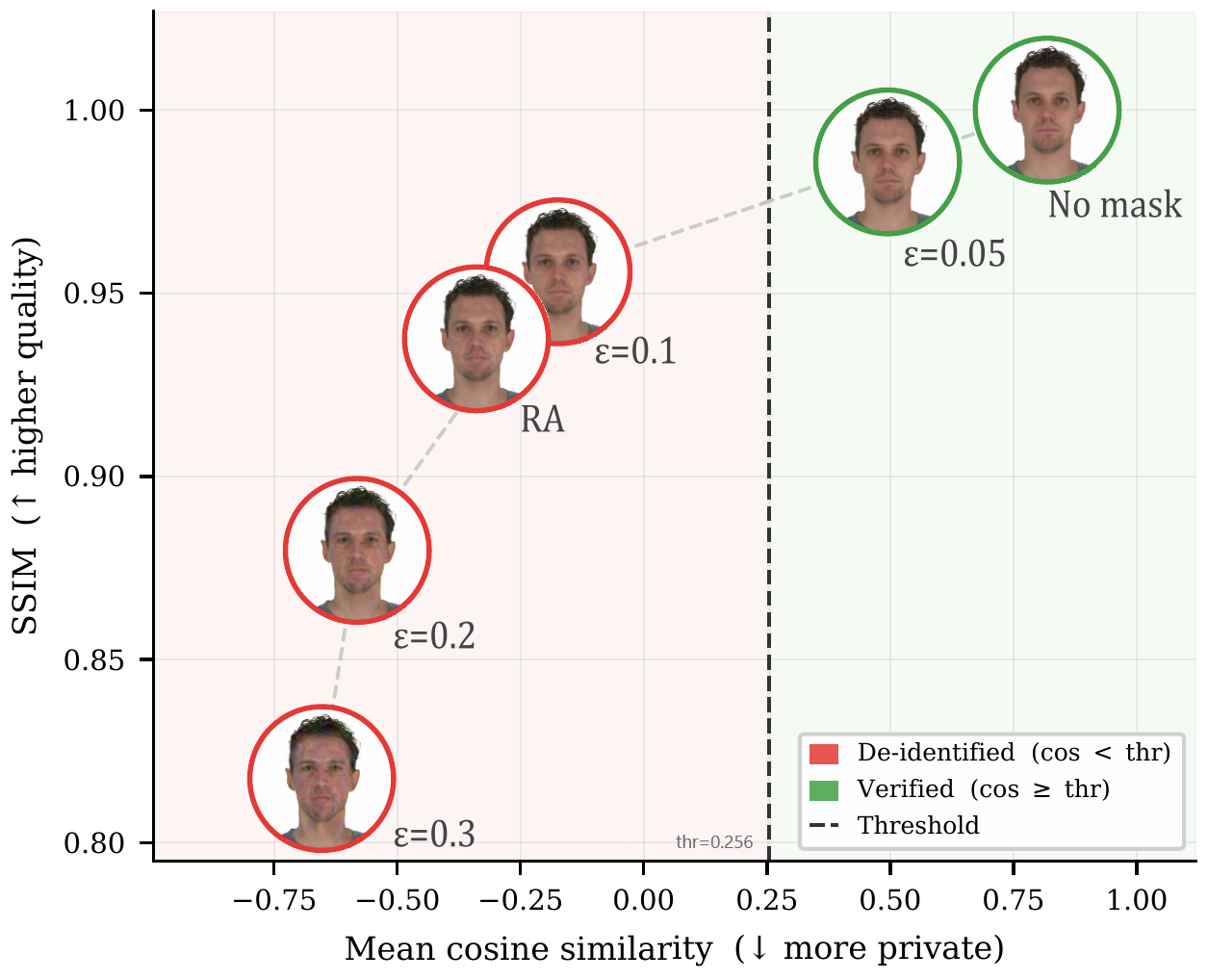}
    \caption{Privacy--quality trade-off for \our{} variants, shown for a~single example subject. Each point plots a~variant's mean cosine similarity to the genuine reference over the $81$ rendered viewpoints of the VRL grid (x-axis, $\downarrow$ more private) against its SSIM (y-axis, $\uparrow$ higher quality), scored by the held-out IR152 recognizer ($\tau^\star{=}0.256$). The dashed line marks the verification threshold, separating de-identified (red, left) from verifiable (green, right) points. \our{} RA achieves the best balance}
    \label{fig:tradeoff}
\end{figure}

Table~\ref{tab:utility} reports utility, while Fig.~\ref{fig:tradeoff} shows the privacy vs quality trade-off across budgets. \our{} RA achieves the strongest balance, combining low leakage with SSIM $0.93$, PSNR $32.7\,\mathrm{dB}$, and FID $29.6$, while baselines approaching its privacy incur substantially greater fidelity losses. Uniform \our{} trades fidelity for increasing privacy as the budget grows, maintaining high gender and emotion agreement. Lower race agreement likely reflects color perturbations that shift skin tone, a~feature apparently used by the race classifier. FaceAnonSimple and differential-privacy baselines target a~different task, replacing the subject's likeness, and are included to situate \our{} in the broader privacy and utility landscape. Nevertheless, \our{} suppresses automated recognition more while preserving visual recognizability.

\paragraph{Commercial API Evaluation}
To assess protection against a~commercial recognition system, we query the Face\textsuperscript{++} Compare API. For each subject, we submit the masked frontal render and the genuine reference to the API endpoint and record the returned confidence score, which ranges from $0$ to $100$, with higher values indicating a~stronger identity match; we report a~verification as positive when the score exceeds the API's recommended threshold at FAR${=}10^{-3}$ ($62.3$). On unmasked avatars, Face\textsuperscript{++} verifies every subject (confidence $93.9{\pm}1.9$, $100\%$ verification). \our{} substantially lowers the match confidence: uniform masking at $\epsilon{=}0.2$ and $\epsilon{=}0.3$ yields confidence $33.0{\pm}9.3$ and $21.9{\pm}7.8$ (both $0\%$ verification), and \our{} RA confidence $46.2{\pm}10.0$ ($3.3\%$). As in Tab.~\ref{tab:privacy}, the smallest budget $\epsilon{=}0.1$ remains insufficient ($67.9{\pm}8.6$, $71.7\%$).

\paragraph{Robustness to Transformations}
To test robustness against protection removal, we evaluate \our{} under $49$ image transformations spanning four groups: (1) standard degradations including JPEG compression, Gaussian and salt-and-pepper noise, blur, resolution downscaling, and color/brightness adjustments, (2) classical adversarial purification including Non-Local Means (NLM) and Total Variation (TV) denoising, wavelet denoising, and iterative JPEG laundering, (3) neural purification including GFPGAN~\cite{gfpgan} face restoration and Real-ESRGAN~\cite{realesrgan} super-resolution, and (4) combined pipelines simulating social-media and messaging compression. We score each transformed render against the held-out recognizer at the same TAR@FAR${=}10^{-3}$ threshold as Tab.~\ref{tab:privacy} and report the mean verification rate over the $60$ subjects (per-transformation results in Tab.~\ref{tab:robustness}). Uniform \our{} ($\epsilon{=}0.2$) is highly robust: mean verification across all transformations is $0.44\%$, with the largest residuals under two purification attacks (NLM and TV denoising, $3.33\%$). Neural restoration methods do not recover identity (GFPGAN and Real-ESRGAN both yield $0\%$ verification). The region-adaptive variant \our{} RA trades some of this robustness for its higher fidelity: mean verification rises to $4.83\%$, with the largest residuals under aggressive purification (full purification pipeline $15.0\%$, $5\%$ salt-and-pepper noise $13.3\%$, GFPGAN and NLM $11.7\%$), consistent with its smaller overall perturbation magnitude, which yields higher fidelity but leaves less signal to survive purification.

\section{Conclusion}
We introduced \our{}, an identity masking method for 3D Gaussian avatars that perturbs only the DC color coefficients using PGD, leaving geometry, view-dependent appearance, and the subject's visual likeness intact. Because the protection is optimized directly into the avatar, it is carried into every rendered viewpoint and animation by design, rather than reapplied per frame as in 2D masking. This lets users share realistic avatars for video calls, social VR, and similar applications while resisting automated verification and identity theft. Our experiments show that these color-only perturbations reduce automated recognition to near-zero while preserving perceptual fidelity and key facial attributes, providing a~practical route to privacy that keeps photorealistic 3D avatars usable for animation and interactive applications.

\noindent\textbf{Limitations and Future Work.} \our{} perturbs only the DC coefficients of spherical harmonics, which preserves geometry but constrains the masking signal, so a~fraction of oblique viewpoints can retain residual identity cues. Future work could address this with a~more focused optimization that concentrates the adversarial signal on identity-discriminative features under an explicit perceptual constraint, suppressing residual leakage without sacrificing fidelity. As \our{} alters only color, its protection is by construction confined to the RGB-based recognition of our assumed threat model; extending it to depth-based geometric verification would require perturbing the geometry itself, which is non-trivial because shape changes must preserve an anatomically plausible face rather than introduce structural artifacts such as displaced or duplicated features.

\section*{Acknowledgements}
The work of A. Pechko and P. Spurek was supported by the project Effective Rendering of 3D Objects Using Gaussian Splatting in an Augmented Reality Environment (FENG.02.02-IP.05- 0114/23), carried out under the First Team programme of the Foundation for Polish Science and co-financed by the European Union through the European Funds for Smart Economy 2021–2027 (FENG). D.~Wolkiewicz and P. Syga were supported by Department of Artificial Intelligence, Wrocław University of Science and Technology.
{
    \small
    \bibliographystyle{ieeenat_fullname}
    \bibliography{references}
}

\clearpage
\appendix
\phantomsection
\addcontentsline{toc}{section}{Appendix}
\section*{Appendix}

\FloatBarrier
\section{Temporal Consistency Under Motion and Expression Changes}
\label{sect:temporal}
We test whether the embedded perturbation stays stable, and identity stays suppressed, while the avatar is animated, simulating real-world use. For 10 subjects, we render both the masked (\our{}, $\epsilon{=}0.2$) and the unmasked avatar over the full recorded expression range (mouth-open, squinting, head-pose extremes) while the camera orbits on a~Lissajous trajectory, so that every consecutive frame pair differs in both viewpoint and expression. To make the comparison identical across subjects, all avatars are driven by a~single shared motion sequence (the FLAME parameters of one recording), while each subject keeps its own identity shape. Identity is scored by the held-out IR152 recognizer.

\begin{figure}[tbp]
  \centering
  \includegraphics[width=1\linewidth]{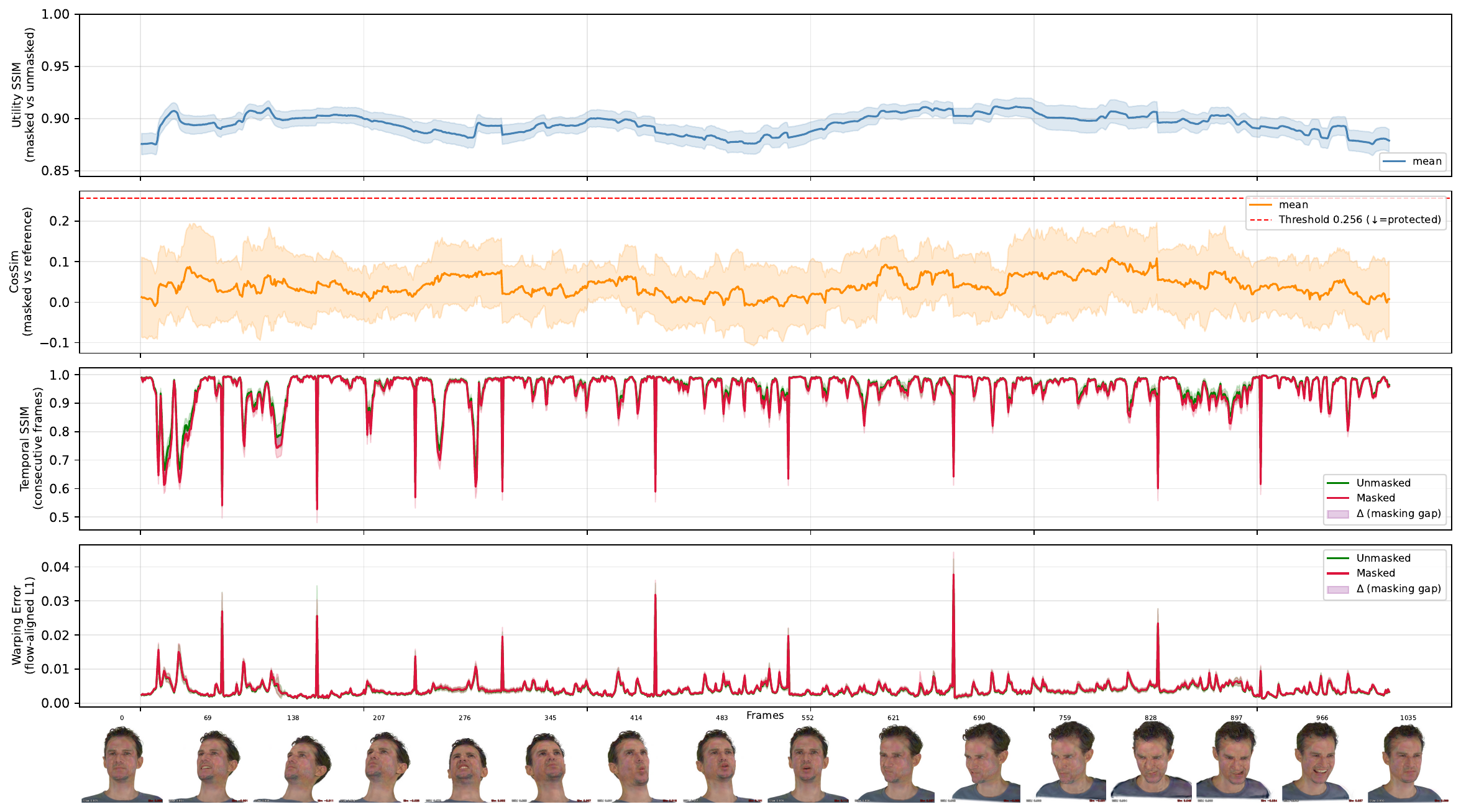}
  \caption{Temporal and privacy-preservation consistency of \our{} ($\epsilon{=}0.2$) masking during motion, spanning expression and viewpoint changes, scored by the held-out IR152 recognizer}
  \label{fig:motion}
\end{figure}

Figure~\ref{fig:motion} reports four per-frame measures, from top to bottom. (1) \emph{Utility SSIM}: the SSIM between the masked and unmasked renders of the same pose, quantifying how much the masking distorts overall image quality at each timestep. (2) \emph{Identity cosine}: the cosine similarity between the masked frame's face embedding and a~fixed reference image of the true identity, measuring whether the masked avatar is recognized as the target frame by frame. (3) \emph{Temporal SSIM}: the SSIM between consecutive frames ($i$ and $i{-}1$), plotted separately for the unmasked and masked sequences with a~shaded region marking the gap between them, quantifying how temporally smooth (low-flicker) each sequence is. (4) \emph{Temporal warping error}~\cite{lai2018learning}: registers each pair of consecutive frames with optical flow (computed with RAFT~\cite{teed2020raft}), warps the previous frame into the current one, and measures the mean residual between the warped and the actual frame, averaged over the sequence; by discounting the motion that the flow already explains, it isolates flicker introduced beyond the underlying head and expression movement. Higher is better for the SSIM measures and lower is better for the warping error. Lines show per-avatar means and bands span $\pm1\sigma$.

Masking introduces essentially no temporal degradation: temporal SSIM is $0.955{\pm}0.060$ unmasked versus $0.948{\pm}0.069$ masked ($\Delta{=}0.007$), and the warping error rises from $3.96{\times}10^{-3}$ to $4.13{\times}10^{-3}$ ($+4.2\%$, within noise). The two temporal curves overlap, with matched expression-driven dips, confirming that these reflect FLAME deformation rather than pattern-induced flicker. Identity protection holds across the full sequence: the mean masked cosine similarity is $0.041{\pm}0.083$, against $0.611$ for the unmasked avatar, well below the IR152 verification threshold ($\tau{=}0.256$). Only $0.38\%$ of the $11{,}180$ rendered frames exceed the threshold; seven of the ten subjects never do, while three show brief exceedances at extreme poses (worst-frame cosine up to $0.335$).

\FloatBarrier
\section{Robustness to Image Transformations}
\label{sect:robustness}
We evaluate the robustness of \our{} (base and RA, $\epsilon=0.2$) against various image transformations spanning four categories: (1) standard degradations: JPEG compression at varying quality levels and iterative JPEG laundering, additive Gaussian and salt-and-pepper noise, Gaussian and motion blur, resolution downsampling and upscaling, color, brightness, and contrast adjustments, and a~random transformation; (2) classical adversarial purification: median and bilateral filtering, Non-Local Means denoising, Total Variation (TV) minimization, wavelet-based denoising, feature squeezing, and bit-depth reduction; (3) neural network-based purification: GFPGAN face restoration~\cite{gfpgan}, which explicitly targets facial artifacts, and Real-ESRGAN super-resolution~\cite{realesrgan}; and (4) combined pipelines simulating social media and instant messaging compression chains.

Table~\ref{tab:robustness} reports verification rates for both \our{} variants across all 49 transformations. Uniform \our{} (base, $\epsilon{=}0.2$) is highly robust, with a~mean verification rate of $0.44\%$ and no transformation exceeding $3.33\%$. The region-adaptive variant (RA), whose smaller perturbation magnitude yields higher fidelity, is correspondingly less robust (mean $4.83\%$), with its largest residuals concentrated under aggressive purification. The following observations emerge per category:

\paragraph{Standard Degradations} Under base masking, all JPEG levels (quality 10--70 and iterative laundering), Gaussian and motion blur, resolution downscaling, and color, brightness, contrast, and gamma adjustments stay at or below $1.67\%$ verification, with most reaching $0\%$. RA retains a~small systematic residual (typically $3.33\%$) across these benign transformations and rises under signal-disrupting noise: salt-and-pepper at $5\%$ density ($13.33\%$), saturation reduction ($11.67\%$), and Gaussian noise $\sigma{=}20$ ($10\%$) are its weakest points, as the added randomness partially erodes the lower-magnitude adversarial signal.

\paragraph{Classical Adversarial Purification} Median filtering, bilateral filtering, feature squeezing, and bit-depth reduction leave base protection essentially intact ($0\%$). Its strongest base residuals come from denoising aimed at adversarial perturbations, namely Non-Local Means (strong) and TV denoising at $3.33\%$, which remain negligible. RA is more exposed: Non-Local Means ($11.67\%$), TV denoising ($10\%$), strong Non-Local Means ($8.33\%$), wavelet denoising ($6.67\%$), and 4-bit quantization ($6.67\%$) produce its highest leakage in this category, consistent with purification eroding the weaker RA signal.

\paragraph{Neural Network Purification} Both GFPGAN~\cite{gfpgan} face restoration and Real-ESRGAN~\cite{realesrgan} super-resolution leave base protection at $0\%$. Notably, GFPGAN visually removes perturbation artifacts yet the reconstructed faces remain unrecognized under base masking, indicating that \our{} encodes identity-relevant shifts in embedding space rather than adding removable structured noise. RA, carrying a~smaller signal, is partially recovered by these neural restorers (GFPGAN $11.67\%$, Real-ESRGAN $5\%$).

\paragraph{Combined Pipelines} Four multi-stage pipelines are evaluated. The \emph{social media} pipeline simulates an upload/download cycle: 0.8$\times$ downscale, JPEG compression (quality~75), upscale to original resolution, and unsharp masking. The \emph{messaging} pipeline applies a~more aggressive variant: 0.5$\times$ downscale, heavy JPEG compression (quality~40), and upscale. The \emph{adversarial purification} pipeline chains a~$3\times3$ median filter, bilateral filtering ($d=9$, $\sigma=75$), and JPEG compression (quality~80). The \emph{full purification} pipeline combines Non-Local Means denoising, bilateral filtering, GFPGAN neural face restoration, and a~final JPEG pass (quality~90). All four pipelines leave base masking at or below $1.67\%$. Under RA the first three pipelines stay at $3.33\%$, while the full purification pipeline, its most aggressive combination, yields RA's highest verification rate overall ($15\%$), reflecting the cumulative effect of stacking denoising, neural restoration, and compression on the lower-magnitude RA signal.

\begin{table}[htp]
\centering
\renewcommand{\arraystretch}{1.05}
{\setlength{\tabcolsep}{6pt}
{\footnotesize
\caption{Verification rate (\%) under image transformations for \our{} base and RA ($\epsilon=0.2$, 60 subjects, scored by the held-out IR152 recognizer at TAR@FAR$=10^{-3}$). Lower is better ($\downarrow$)}
\label{tab:robustness}
\begin{tabular}{llrr}
\hline
\textbf{Category} & \textbf{Transformation} & \makecell{\textbf{base}\\\textbf{(\%)$\downarrow$}} & \makecell{\textbf{RA}\\\textbf{(\%)$\downarrow$}} \\ \hline
\multirow{31}{*}{Standard degradations}
 & JPEG quality 70             & 0    & 3.33 \\
 & JPEG quality 50             & 0    & 3.33 \\
 & JPEG quality 30             & 0    & 3.33 \\
 & JPEG quality 10             & 1.67 & 5.00 \\
 & Iterative JPEG (5$\times$)  & 0    & 3.33 \\
 & Iterative JPEG (10$\times$) & 0    & 3.33 \\
 & Gaussian noise $\sigma=5$   & 0    & 3.33 \\
 & Gaussian noise $\sigma=10$  & 1.67 & 3.33 \\
 & Gaussian noise $\sigma=20$  & 1.67 & 10.00 \\
 & Salt-and-pepper 1\%         & 0    & 6.67 \\
 & Salt-and-pepper 5\%         & 1.67 & 13.33 \\
 & Gaussian blur $3\times3$    & 0    & 3.33 \\
 & Gaussian blur $5\times5$    & 0    & 3.33 \\
 & Gaussian blur $7\times7$    & 0    & 3.33 \\
 & Motion blur                 & 0    & 3.33 \\
 & Blur-sharpen ($3\times3$)   & 0    & 1.67 \\
 & Blur-sharpen ($5\times5$)   & 0    & 0 \\
 & Downscale $\times0.5$       & 0    & 3.33 \\
 & Downscale $\times0.25$      & 0    & 3.33 \\
 & Brightness $+30$            & 0    & 3.33 \\
 & Brightness $-30$            & 0    & 3.33 \\
 & Contrast $0.7\times$        & 0    & 3.33 \\
 & Contrast $1.3\times$        & 0    & 3.33 \\
 & Gamma $0.7$                 & 0    & 3.33 \\
 & Gamma $1.5$                 & 0    & 3.33 \\
 & Saturation $0.5\times$      & 1.67 & 11.67 \\
 & Saturation $1.5\times$      & 0    & 1.67 \\
 & Histogram equalization      & 0    & 1.67 \\
 & CLAHE                       & 0    & 3.33 \\
 & Sharpen                     & 1.67 & 3.33 \\
 & Random transform            & 0    & 3.33 \\
\hline
\multirow{12}{*}{Classical purification}
 & Median $3\times3$           & 0    & 3.33 \\
 & Median $5\times5$           & 0    & 3.33 \\
 & Non-Local Means             & 1.67 & 11.67 \\
 & Non-Local Means (strong)    & 3.33 & 8.33 \\
 & Wavelet denoising           & 0    & 6.67 \\
 & Wavelet denoising (strong)  & 1.67 & 6.67 \\
 & TV denoising                & 3.33 & 10.00 \\
 & TV denoising (strong)       & 0    & 3.33 \\
 & Bilateral filter            & 0    & 3.33 \\
 & Feature squeezing           & 0    & 3.33 \\
 & Bit-depth 4-bit             & 0    & 6.67 \\
 & Bit-depth 6-bit             & 0    & 3.33 \\
\hline
\multirow{2}{*}{Neural purification}
 & GFPGAN face restoration     & 0    & 11.67 \\
 & Real-ESRGAN super-resolution & 0   & 5.00 \\
\hline
\multirow{4}{*}{Combined pipelines}
 & Social media pipeline       & 0    & 3.33 \\
 & Messaging pipeline          & 0    & 3.33 \\
 & Adversarial purif.\ pipeline & 0   & 3.33 \\
 & Full pipeline               & 1.67 & 15.00 \\
\hline
\end{tabular}
}
}
\end{table}

\FloatBarrier
\section{Region-Adaptive Perturbation Budget: Multiplier Selection}
\label{sect:ra_sweep}
Our region-adaptive variant (\our{} RA) assigns per-region $\epsilon$ multipliers to the base budget $\epsilon=0.2$, concentrating perturbations in identity-critical areas while reducing visible artifacts in smooth, perceptually sensitive regions. To determine the final multipliers reported in the main text, we conducted a~preliminary sweep over six candidate configurations, evaluated on 10 subjects from the NeRSemble dataset.

Table~\ref{tab:ra_sweep_configs} lists the per-region multipliers for each candidate, and Table~\ref{tab:ra_sweep_results} reports the corresponding quantitative results.

\begin{table}[htp]
\centering
\renewcommand{\arraystretch}{1.1}
{\setlength{\tabcolsep}{3.5pt}
{\footnotesize
\caption{Per-region $\epsilon$ multiplier configurations evaluated in the sweep. Region groups: \textbf{Identity-critical} (eyes, nose), \textbf{Balanced} (lips), \textbf{Smooth} (skin, forehead, ears, scalp). Unspecified regions inherit the base multiplier of $1.0$}
\label{tab:ra_sweep_configs}
\begin{tabular}{llccccccc}
\hline
\textbf{Config} & \textbf{Description} & \textbf{Eyes} & \textbf{Nose} & \textbf{Lips} & \textbf{Skin} & \textbf{Forehead} & \textbf{Ears} & \textbf{Scalp} \\ \hline
00 Uniform         & No adaptation                        & 1.0 & 1.0 & 1.0 & 1.0 & 1.0 & 1.0 & 1.0 \\
01 Final RA*     & Identity boost, smooth reduction     & 1.3 & 1.2 & 1.0 & 0.7 & 0.6 & 0.7 & 0.4 \\
02 Boost Eye+Nose  & Stronger boost on top-2 regions      & 1.5 & 1.4 & 1.0 & 0.7 & 0.6 & 0.7 & 0.4 \\
03 Boost E+N+Lips  & Include lips as identity-critical    & 1.5 & 1.4 & 1.3 & 0.7 & 0.6 & 0.7 & 0.4 \\
04 Two-tier        & Binary high/low allocation           & 1.4 & 1.4 & 1.4 & 0.4 & 0.4 & 0.4 & 0.4 \\
05 Tight Smooth    & Aggressive smooth-region reduction   & 1.3 & 1.2 & 1.0 & 0.5 & 0.4 & 0.5 & 0.2 \\ \hline
\multicolumn{9}{l}{*\,Configuration adopted in the main text.}
\end{tabular}
}
}
\end{table}

\begin{table}[htp]
\centering
\renewcommand{\arraystretch}{1.1}
{\setlength{\tabcolsep}{6pt}
{\small
\caption{Quantitative results for each multiplier configuration, evaluated on 10 NeRSemble subjects (preliminary). Verification rate is the fraction of subjects verified above the TAR@FAR threshold ($\tau=0.256$). Lower Avg.\ Sim.\ provides a~larger safety margin from the threshold}
\label{tab:ra_sweep_results}
\begin{tabular}{lrrrr}
\hline
\textbf{Config} & \makecell{\textbf{Verification}\\\textbf{Rate (\%) $\downarrow$}} & \textbf{SSIM $\uparrow$} & \makecell{\textbf{PSNR}\\\textbf{(dB) $\uparrow$}} & \makecell{\textbf{Avg.}\\\textbf{Sim. $\downarrow$}} \\ \hline
00 Uniform        & 0   & 0.899 & 31.05 & $-0.259$ \\
\rowcolor{gray!12}01 Final RA   & 0   & 0.937 & 33.74 & $-0.061$ \\
02 Boost Eye+Nose & 20   & 0.950 & 34.99 & $+0.167$ \\
03 Boost E+N+Lips & 0   & 0.949 & 34.95 & $+0.134$ \\
04 Two-tier       & 0   & 0.953 & 35.26 & $+0.182$ \\
05 Tight Smooth   & 30  & 0.960 & 36.14 & $+0.207$ \\ \hline
\end{tabular}
}
}
\end{table}

While configurations 02--05 achieve better SSIM and PSNR than the chosen config 01, they do so at the cost of a~reduced safety margin: their average cosine similarities are positive and approach the verification threshold. Configurations 02 and 05 already leak identity on the preliminary subjects. The chosen configuration (01 Final RA) provides a~margin below the threshold (avg.\ sim.\ $=-0.061$) while substantially improving visual quality over the uniform baseline (SSIM $0.937$ vs.\ $0.899$). The uniform config achieves the strongest identity suppression (avg.\ sim.\ $=-0.259$) but does so at the cost of visual quality.

\FloatBarrier
\section{Attacks on Different Gaussian Avatar Components}
We investigate the effect of adversarial perturbations applied to components of the 3D Gaussian Splatting avatar representation beyond the DC (base color) coefficients. Specifically, we evaluate attacks on: (1) AC coefficients (higher-order spherical harmonics), (2) Gaussian positions, (3) scale, (4) rotation, and (5) opacity. These experiments were conducted as a~preliminary study on 10 subjects from the NeRSemble dataset.

\begin{figure}[tbp]
    \centering
    \includegraphics[width=1\linewidth]{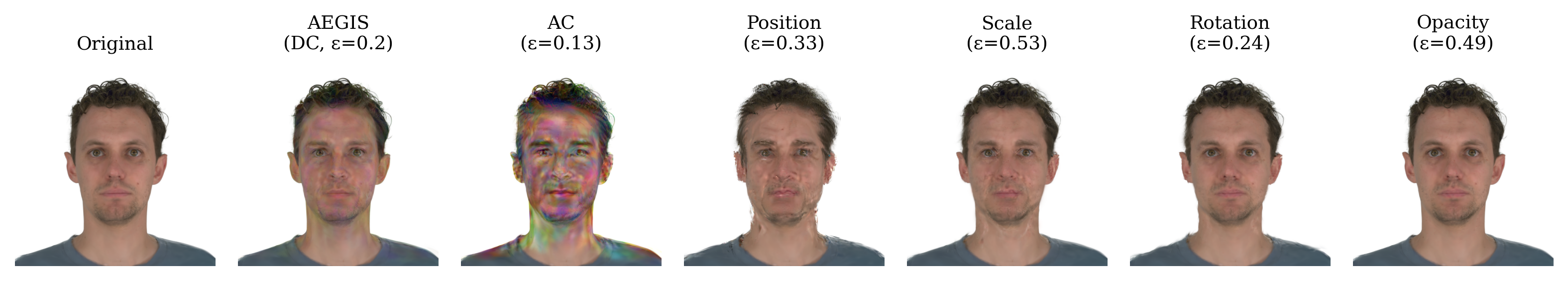}
    \caption{Qualitative comparison of adversarial perturbations applied to different components of the 3D Gaussian avatar representation. Each column shows the masked render under an attack targeting a~single component, alongside the original and \our{}'s DC-based masking}
    \label{fig:diff_comps}
\end{figure}

For each component we run the same $\ell_\infty$-bounded PGD as \our{} ($200$ steps, the three-surrogate ensemble with IR152 held out for scoring), keeping the evaluation protocol identical to the DC-based attack. Because the components occupy different parameter spaces with very different dynamic ranges, a~single shared $\epsilon$ would be an unfair comparison; we therefore assign each component a~range-matched budget:
\begin{equation}
    \epsilon_c = \epsilon_{\mathrm{DC}}\cdot \frac{1}{S}\sum_{s=1}^{S} \frac{R_c^{(s)}}{R_{\mathrm{DC}}^{(s)}}~,
\end{equation}
where $R^{(s)}$ is the dynamic range of the corresponding tensor for subject $s$, $S$ the number of subjects, and $\epsilon_{\mathrm{DC}}{=}0.2$. Averaged over the $10$ subjects this yields $\epsilon$ of $0.13$ (AC), $0.33$ (position), $0.53$ (scale), $0.24$ (rotation), and $0.49$ (opacity). Results are summarized in Table~\ref{tab:diff_comps}, with qualitative comparisons shown in Figure~\ref{fig:diff_comps}.

\begin{table}[htp]
\centering
\caption{Quantitative evaluation of adversarial attacks applied to different components of the 3D Gaussian avatar representation, compared to \our{}'s DC-based configuration (shaded). Preliminary study on 10 NeRSemble subjects under the ensemble surrogate protocol with IR152 held out. Lower is better for Rank-$k$/Verification/FID/Age; higher for SSIM/PSNR/Race/Gender/Emotion}
\label{tab:diff_comps}
\renewcommand{\arraystretch}{1}
{\setlength{\tabcolsep}{7pt}
{\fontsize{9pt}{11pt}\selectfont
\resizebox{\textwidth}{!}{
    \begin{tabular}{lrrrrrrrrrr}
    \hline
    \bf Method & \makecell{\textbf{Rank}\\\textbf{1 (\%) $\downarrow$}} & \makecell{\textbf{Rank}\\\textbf{50 (\%) $\downarrow$}} & \makecell{\textbf{Verification}\\\textbf{(\%) $\downarrow$}} & \bf SSIM $\uparrow$ & \makecell{\textbf{PSNR}\\\textbf{(dB) $\uparrow$}} & \bf FID $\downarrow$ & \makecell{\textbf{Age}\\\textbf{(diff.) $\downarrow$}} & \makecell{\textbf{Race}\\\textbf{(\%) $\uparrow$}} & \makecell{\textbf{Gender}\\\textbf{(\%) $\uparrow$}} & \makecell{\textbf{Emotion}\\\textbf{(\%) $\uparrow$}} \\ \hline
    \rowcolor{gray!12}\bf \our{} (DC) & 0 & 0 & 0 & 0.878 & 29.35 & 64.16 & 4.80 & 30 & 90 & 80 \\
    \hline
    AC & 0 & 0 & 0 & 0.584 & 20.65 & 141.37 & 8.80 & 10 & 90 & 30 \\
    Opacity & 100 & 100 & 100 & 0.981 & 38.91 & 11.54 & 2.20 & 80 & 90 & 90 \\
    Position & 0 & 0 & 0 & 0.711 & 23.59 & 109.24 & 9.40 & 30 & 90 & 40 \\
    Rotation & 50 & 90 & 80 & 0.923 & 30.23 & 39.97 & 4.00 & 70 & 90 & 90 \\
    Scale & 10 & 20 & 10 & 0.894 & 29.23 & 48.45 & 6.80 & 30 & 90 & 80 \\
    \hline
    \end{tabular}
}
}
}
\end{table}

The non-DC components fall into two failure modes. Perturbing opacity or rotation barely affects the rendered identity and the held-out recognizer still verifies $100\%$ and $80\%$ of avatars, respectively. Scale only partially de-identifies ($10\%$ verification, $20\%$ Rank-50). Perturbing AC coefficients or Gaussian positions does suppress identity ($0\%$ verification) but corrupts the view-dependent appearance or geometry, severely degrading perceptual quality (AC: SSIM $0.584$, FID $141.4$; position: SSIM $0.711$, FID $109.2$).

DC perturbation achieves the best trade-off between our two objectives. The components that best preserve fidelity (opacity, rotation, and scale) fail to fully suppress identity, whereas the two that reach $0\%$ verification (AC and position) substantially degrade quality. Perturbing the DC coefficients is the only configuration that combines strong de-identification with high fidelity.

\FloatBarrier
\section{Different Adversarial Optimization Strategies}
We extend our ablation by evaluating a~broader set of adversarial optimization strategies, including both $\ell_\infty$ and Euclidean ($\ell_2$) perturbation regimes. Beyond the iterative PGD attack used in \our{}, we consider two additional families of methods. First, we apply the Fast Gradient Sign Method (FGSM)~\cite{goodfellow2014explaining}, a~single-step linearization that perturbs parameters in the direction of the gradient sign. Second, we adopt the Decoupled Direction and Norm (DDN) attack~\cite{rony2019decoupling}, which iteratively optimizes perturbations by separating the gradient direction from the update magnitude, thereby seeking the minimal $\ell_2$ norm necessary to induce a~successful attack. These experiments were conducted as a~preliminary study on 10 subjects from the NeRSemble dataset.

\begin{figure}[tbp]
    \centering
    \includegraphics[width=1\linewidth]{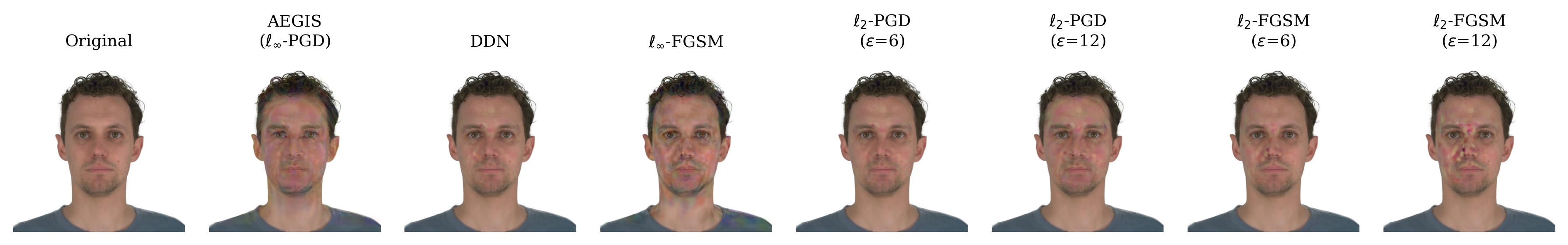}
    \caption{Qualitative comparison of identity masking under different adversarial optimization strategies (DDN, FGSM, and PGD under $\ell_\infty$ and $\ell_2$ norms), all targeting DC and shown alongside the original render and \our{}'s $\ell_\infty$-PGD masking}
    \label{fig:diff_attacks}
\end{figure}

In the high-dimensional DC coefficient space, directly converting our $\ell_\infty$ budget ($\epsilon_\infty = 0.2$) into its $\ell_2$ equivalent yields $\epsilon_2 = \epsilon_\infty \sqrt{d}$, where $d$ is the dimensionality of the masked DC tensor. With $d \approx 3.4{\times}10^{5}$ this gives $\epsilon_2 \approx 116$, a~magnitude that introduces severe, visually unacceptable distortions. To obtain a~realistic scale we use DDN to estimate the minimal effective $\ell_2$ norm for a~successful attack; it converges to $\|\cdot\|_2 \approx 5.7$ on average over the 10 subjects. We therefore evaluate fixed-budget $\ell_2$ attacks at $\epsilon_2 = 6$ (near this minimum) and $\epsilon_2 = 12$ (roughly twice it). Quantitative comparisons for all $\ell_\infty$ and $\ell_2$ strategies are presented in Table~\ref{tab:diff_attacks}, with qualitative visualizations in Figure~\ref{fig:diff_attacks}.

\begin{table}[htp]
\centering
\caption{Quantitative comparison of adversarial optimization strategies (all targeting DC) under $\ell_\infty$ and $\ell_2$ constraints, compared to \our{}'s $\ell_\infty$-PGD configuration (shaded). Preliminary study on 10 NeRSemble subjects under the ensemble surrogate protocol with IR152 held out. Lower is better for Rank-$k$/Verification/FID/Age; higher for SSIM/PSNR/Race/Gender/Emotion}
\label{tab:diff_attacks}
\renewcommand{\arraystretch}{1}
{\setlength{\tabcolsep}{7pt}
{\fontsize{9pt}{11pt}\selectfont
\resizebox{\textwidth}{!}{
    \begin{tabular}{lrrrrrrrrrr}
    \hline
    \bf Method & \makecell{\textbf{Rank}\\\textbf{1 (\%) $\downarrow$}} & \makecell{\textbf{Rank}\\\textbf{50 (\%) $\downarrow$}} & \makecell{\textbf{Verification}\\\textbf{(\%) $\downarrow$}} & \bf SSIM $\uparrow$ & \makecell{\textbf{PSNR}\\\textbf{(dB) $\uparrow$}} & \bf FID $\downarrow$ & \makecell{\textbf{Age}\\\textbf{(diff.) $\downarrow$}} & \makecell{\textbf{Race}\\\textbf{(\%) $\uparrow$}} & \makecell{\textbf{Gender}\\\textbf{(\%) $\uparrow$}} & \makecell{\textbf{Emotion}\\\textbf{(\%) $\uparrow$}} \\ \hline
    \rowcolor{gray!12}\bf \our{} ($\ell_\infty$-PGD) & 0 & 0 & 0 & 0.878 & 29.35 & 64.16 & 4.80 & 30 & 90 & 80 \\
    \hline
    DDN & 80 & 100 & 90 & 0.977 & 38.36 & 12.39 & 4.00 & 40 & 90 & 100 \\
    $\ell_\infty$-FGSM ($\epsilon{=}0.2$) & 60 & 90 & 80 & 0.858 & 29.46 & 74.29 & 7.70 & 50 & 90 & 60 \\
    $\ell_2$-PGD ($\epsilon{=}6$) & 70 & 90 & 80 & 0.975 & 37.76 & 12.94 & 4.50 & 50 & 90 & 90 \\
    $\ell_2$-PGD ($\epsilon{=}12$) & 0 & 20 & 0 & 0.935 & 32.44 & 30.06 & 4.90 & 30 & 90 & 70 \\
    $\ell_2$-FGSM ($\epsilon{=}6$) & 100 & 100 & 100 & 0.974 & 37.63 & 13.61 & 2.10 & 80 & 90 & 90 \\
    $\ell_2$-FGSM ($\epsilon{=}12$) & 100 & 100 & 100 & 0.928 & 31.69 & 43.59 & 6.80 & 50 & 90 & 40 \\
    \hline
    \end{tabular}
}
}
}
\end{table}

In our experiments, breaking black-box verification by perturbing the DC coefficients appears to require two conditions: the attack must be iterative, and it must use a~sufficient perturbation budget. Single-step FGSM is ineffective under both norms (verification between $80$ and $100\%$), and the minimum-norm DDN attack ($\|\cdot\|_2\approx5.7$) leaves $90\%$ verification, and low-budget $\ell_2$-PGD ($\epsilon{=}6$) leaves $80\%$. Among the settings we evaluate, only two de-identify subjects: \our{}'s $\ell_\infty$-PGD and $\ell_2$-PGD at the larger radius $\epsilon{=}12$, both reaching $0\%$ verification. The two lie close on the privacy/fidelity trade-off, and $\ell_2$-PGD ($\epsilon{=}12$) attains somewhat higher reconstruction fidelity (SSIM $0.935$, FID $30.1$) than $\ell_\infty$-PGD (SSIM $0.878$, FID $64.2$); however, it still leaks at retrieval ($20\%$ Rank-50, compared with $0\%$ for $\ell_\infty$-PGD). \our{} adopts the $\ell_\infty$ formulation for two reasons: it achieves more complete suppression, and its per-channel budget bounds the worst-case deviation of every color value, a~guarantee the aggregate $\ell_2$ norm does not provide.

\FloatBarrier
\section{Additional Samples Visualization}
\begin{figure}[tbp]
    \centering
    \includegraphics[width=\linewidth]{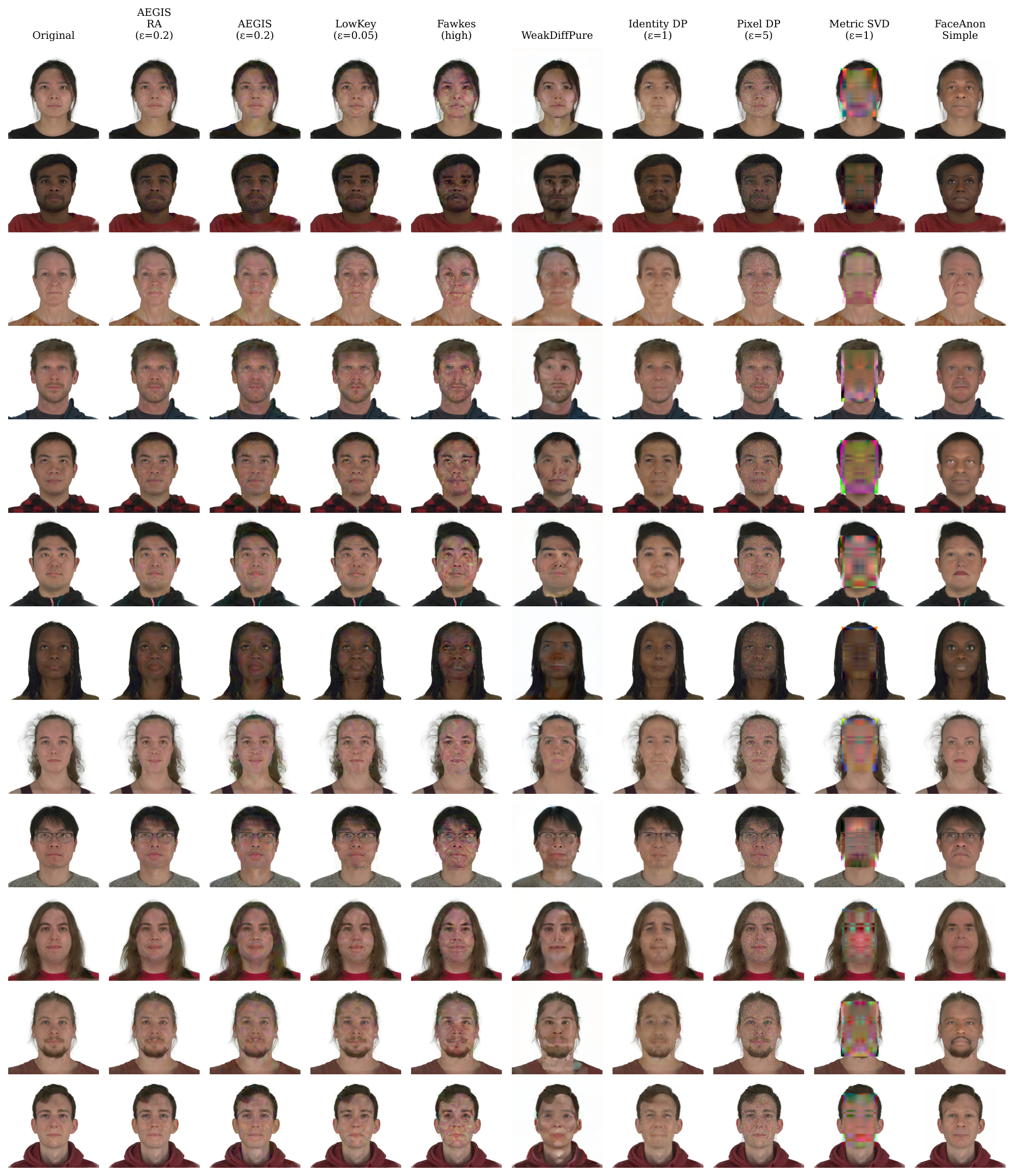}
    \caption{Extended qualitative visualization of avatar masking results for methods evaluated in the main text}
    \label{fig:samples_full}
\end{figure}

We include in Figure~\ref{fig:samples_full} an expanded set of avatar renderings for methods evaluated in the main text. These qualitative examples illustrate the characteristic visual behavior of each approach across a~diverse set of subjects, complementing the quantitative results.

\end{document}